\documentclass[nohyperref]{article}

\usepackage{microtype}
\usepackage{graphicx}
\usepackage{subfigure}
\usepackage{booktabs} 

\usepackage{hyperref}

\usepackage[accepted]{icml2022_pods}

\usepackage{amsmath}
\usepackage{amssymb}
\usepackage{mathtools}
\usepackage{amsthm}

\usepackage[capitalize,noabbrev]{cleveref}

\theoremstyle{plain}

\theoremstyle{definition}

\theoremstyle{remark}

\usepackage[textsize=tiny]{todonotes}
\usepackage{amsmath,bm,graphicx,psfrag,epsf,enumerate,placeins,float,sectsty,amstext,relsize,wrapfig,listings,subfigure,amsfonts,amsthm,mathabx,enumitem,hyperref,grffile,colortbl,dsfont,bbm,setspace, float, xcolor}

\newcommand{\beginsupplement}{ 
        \setcounter{section}{0}
        \renewcommand{\thesection}{S\arabic{section}} %
         \renewcommand{\thesubsection}{\thesection.\arabic{subsection}}
        \setcounter{table}{0}
        \renewcommand{\thetable}{S\arabic{table}} %
        \setcounter{figure}{0}
        \renewcommand{\thefigure}{S\arabic{figure}} %
     }

\usepackage{balance}

\usepackage{dblfloatfix} 

\usepackage{flushend}
\DeclareMathOperator*{\argmax}{argmax}

\allowdisplaybreaks  

\definecolor{LightGray}{gray}{0.9}

\definecolor{lighter-gray}{gray}{0.95} 

\usepackage{xcolor}
 
\definecolor{codegreen}{rgb}{0,0.6,0}
\definecolor{codegray}{rgb}{0.5,0.5,0.5}
\definecolor{codepurple}{rgb}{0.58,0,0.82}
\definecolor{backcolour}{rgb}{0.95,0.95,0.92}

\newcounter{daggerfootnote}

\newcommand{\todored}[1]{\textcolor{red}{TODO: #1}}  

\newcommand{\papertitle}{Back to the Basics: Revisiting  Out-of-Distribution Detection Baselines}

\icmltitlerunning{\papertitle}

\newcommand{\gitlink}{\url{https://github.com/cleanlab/ood-detection-benchmarks}}

\begin{document}

\twocolumn[
\icmltitle{\papertitle}
\icmlsetsymbol{equal}{*}

\begin{icmlauthorlist}
\icmlauthor{Johnson Kuan}{comp}
\icmlauthor{Jonas Mueller}{comp}
\end{icmlauthorlist}

\icmlaffiliation{comp}{Cleanlab}
\icmlcorrespondingauthor{Jonas M}{jonas@cleanlab.ai}

\icmlkeywords{Machine Learning, ICML}

\vskip 0.3in
]



\printAffiliationsAndNotice{}  

\begin{abstract}
We study simple methods for out-of-distribution (OOD) image detection that are compatible with any already trained classifier, relying on only its predictions or learned representations. 
Evaluating the OOD detection performance of various methods when utilized with ResNet-50 and Swin Transformer models, we find methods that solely consider the model's predictions can be easily outperformed by also considering the learned representations. Based on our analysis, we advocate for a dead-simple approach that has been neglected in other studies: simply flag as OOD images whose average distance to their $K$ nearest neighbors is large (in the representation space of an image classifier trained on the in-distribution data).
\end{abstract}

\vspace*{-.5em}
\section{Introduction}
\label{sec:intro}
\vspace*{-0.1em}

\emph{Reliability} is the Achilles' heel of many machine learning (ML) systems deployed today. While ML predictions are accurate for inputs that resemble examples previously encountered during model training, these predictions are often completely unreliable for \emph{out-of-distribution} inputs that stem from a different distribution than the one underlying the training samples. While it may be impossible to ensure accurate predictions for inputs with no similar training examples, a \emph{reliable} ML system should at least recognize it is highly uncertain when encountering an OOD input during deployment (for which its prediction will be wildly extrapolated and hence untrustworthy). 

Here we consider a standard setting for OOD detection with data that is being used for classification tasks \cite{hendrycks2016baseline, yang2021generalized}. Given an example with feature values $x$ and class label $y$, OOD detection methods aim to quantify the likelihood that this example stems from the (unknown) data-generating distribution $\mathcal{P}_{XY}$ whose samples form the training dataset $\{(x_i, y_i)\}_{i=1}^n$ vs.\ stemming from a different distribution. Often the label of this example $y$ will be unavailable, and thus broadly applicable OOD scores only depend on $x$ (evaluating it against the unknown training feature distribution $\mathcal{P}_X$).
While many other OOD settings have been considered in which various additional information is available \cite{hendrycks2019exposure, chen2021robust, fort2021exploring}, the straightforward setting outlined above is appealing for its broad applicability.

Related to tasks like \emph{outlier/anomaly/novelty detection}, \emph{open set recognition}, and \emph{uncertainty estimation}, OOD detection was for a long time approached via density estimation and generative modeling \cite{zhang2021understanding}. 
As deep learning became dominant for image data, increasingly sophisticated neural generative models were widely proposed for OOD as were related models like autoencoders \cite{yang2021generalized}. However such OOD detection methods are highly-model specific and these models often require cumbersome tuning (of architectures/hyperparameters) to fit particular data well. While doable in benchmark studies, tuning models for OOD is nontrivial in practical applications that typically lack ground truth regarding which examples are OOD.

In contrast, models can be easily tuned for supervised learning where predictive accuracy can be assessed against ground truth labels. This has lead to massive practical advances in image classification models \cite{beyer2020we}.
Here we consider OOD detection that leverages any trained classification model, regardless how it was trained\footnote{The best OOD methods from our benchmark are available at: \url{https://github.com/cleanlab/cleanlab}}.
As classification performance continues to improve over time, the performance of such model-agnostic OOD methods automatically improves as well \cite{Vaze2022good}, and they can be utilized with model ensembles that better estimate uncertainty \cite{jain2020maximizing}.  

\vspace*{-0.5em}
\section{Methods}
\label{sec:methods}
\vspace*{-0.5em}

Given an image $x$, a trained (neural) classifier outputs predicted class-probabilities $p = h(g(x))$ based on learned intermediate representations (i.e.\ embedding vectors): $z = g(x)$. This work considers OOD methods that solely depend on $p$ or $z$. For example, one can compute an OOD score for $x$ via the \textbf{Maximum Softmax Probability} (MSP): $\argmax_j \{ p_j\}$, or via the \textbf{Entropy}: $-\sum_j p_j \log p_j$.

\citet{hendrycks2016baseline, Vaze2022good} report that the maximum softmax probability with a standard multiclass classifier is effective for OOD image detection compared to more sophisticated models explicitly designed for this purpose. Beyond the simplicity of the approach, this is surprising because MSP with most classifiers (such as neural networks trained via empirical risk minimization) does not quantify the epistemic model uncertainty, and instead estimates the irreducible noise in the true underlying $p(y | x)$ relationship between features and labels (i.e. aleatoric uncertainty). Conventional wisdom says that epistemic uncertainty should be a more appropriate score for out-of-distribution detection, since aleatoric uncertainty measures an example's proximity to the classification decision boundary rather than its ``outlyingness'' \cite{kirsch2021pitfalls}.

Since this surprising discovery, many subtle criticisms have been proposed for OOD detection via generative modeling \cite{nalisnick2018deep,ren2019likelihood}. However generative models feel conceptually more ``right'' for OOD detection than aleatoric uncertainty estimators like MSP. Generative models will flag as OOD examples that are unlikely to have arisen from the distribution underlying the training data. Discriminative neural classifiers can wildly extrapolate in these training data scarce regions of the feature space, producing overconfident predictions that do not flag images in such regions as OOD \cite{kirsch2021pitfalls}.

We hypothesize that the dataset-specific nature and nontrivial training requirements of sophisticated generative models has hampered their practical OOD performance \cite{zhang2021understanding}, and consider a very crude simplification. Rather than a complex density estimator (e.g.\ normalizing flows), GAN, diffusion, or autoencoder architecture, we can simply leverage the $K$ nearest neighbors of any example $x$ and their distance to $x$ as a crude estimate of the local density around $x$ \cite{zhao2020analysis}. In practice, we measure distances between images more meaningfully in the embedding space of a neural network rather than in the original pixel space. Here we simply consider this neural network to be an arbitrary model for image classification that has been fit to the training data. We follow the common choice of cosine distance as an effective similarity metric for learned image representations \cite{chen2020simple}.

To score how OOD a given image $x$ is, we take the \emph{average distance} between $x$ and each of its $K$ nearest neighbors (KNN) from the training data \cite{angiulli2002fast,luan2021out,ghosh2022knn,bergman2020deep}. 
This approach, which we call the \textbf{KNN Distance}, is a natural strategy for OOD detection; we are unsure why it has been overlooked in recent OOD image detection studies.
Computation of KNN Distance can be accelerated via a multitude of approximate KNN algorithms \cite{aumuller2017ann}.

Another crude estimate for the local density around $x$ (based on Gaussian mixture approximation instead of nonparametric nearest neighbors) can be obtained via the \textbf{Mahalanobis} Distance, which has recently become  popular for OOD image detection \cite{lee2018simple, fort2021exploring}. 
Like KNN Distance, we compute Mahalanobis Distance in the embedding space of a trained neural image classifier. 
\textbf{Relative Mahalanobis Distance} (RMD) aims to improve upon the Mahalanobis Distance by addressing  recently hypothesized flaws of density-based OOD detection \cite{ren2021simplefix}.

\section{Experiments}
\label{sec:experiments}
\vspace*{-0.4em}

We consider the general setting in which during training, OOD methods only have access to the training data, and during testing, one unlabeled test example is presented at a time which should be given an OOD score. 
We evaluate various OOD detection methods under the standard paradigm considered in existing OOD benchmarks\footnote{Our experiments can be reproduced at: \gitlink{}}, in which examples from two different existing datasets are assessed as either being from the same dataset as the training data (in-distribution) or a different dataset (out-of-distribution). 

We first fit our neural network to the \emph{training} set of the in-distribution dataset and use it to obtain embeddings and predictions from the training data. Subsequently, we consider examples from the \emph{test set} of either: the same dataset or the different out-of-distribution dataset. Each test example is passed through the neural network to obtain an embedding and prediction, which can then be contrasted against the corresponding training data quantities to determine if this example is out-of-distribution. Each  model-agnostic OOD detection method considered here quantifies this contrast via a different statistic. To gauge the precision/recall of each OOD method, we study AUROC curves that compare the values of these statistics against which test examples really were out-of-distribution. 

\textbf{Datasets.} \ 
We consider various datasets as in-distribution vs.\ OOD pairs. For each pair of datasets, we run one experiment where the second dataset is considered OOD and a symmetric second experiment where their roles are swapped such that the first dataset is OOD. We evaluate on datasets with color images: \textbf{cifar-10} \& \textbf{cifar-100} \cite{cifar}, and those with grayscale images: \textbf{mnist} \cite{deng2012mnist}, \textbf{fashion-mnist} \cite{xiao2017fashion}, \textbf{roman-numeral} \cite{competition}.

\textbf{Models.} \ 
We trained two image classification models on each in-distribution dataset: \textbf{ResNet-50}  \cite{he2016deep} and \textbf{Swin Transformer} \cite{liu2021swin}.
Both models were tuned and fit using the \texttt{autogluon} AutoML package \cite{agtabular}.
The Swin Transformer is a slightly higher accuracy classifier than ResNet-50 on each of the in-distribution test sets (Table \ref{tab:acc}). This agrees with the findings of previous work \cite{fort2021exploring,Vaze2022good} that indicates we should  prefer to use Transformer architectures for OOD image detection in today's practical applications. 
Thus we give more weight to the conclusions drawn for this model when interpreting results below.

The dimensionality of the image embeddings $z$ (i.e.\ penultimate layer representations) obtained from each model are: 1024 for Swin Transformer, 2048 for ResNet-50. 
For all datasets and models, the dimensionality of the predicted class probabilities $p$ is 10, except it is 100 when cifar-100 is used as the in-distribution dataset. 

\textbf{Alternative Methods.} \ 
Recall the KNN Distance only considers the learned model embeddings $z$ and not the model predicted class probabilities $p$. To determine if these predictions provide additional value for KNN-based OOD detection, we consider two alternative variants: \textbf{KNN DistPred} is the same as the KNN Distance method, except we compute distance in a higher-dimensional vector space in which the image embeddings $z$ are concatenated with the predicted probabilities $p$. \textbf{KNN Prediction} is an OOD score based on the cross-entropy between the predicted class probabilities for image $x$ and the \emph{average} predicted class probability vector amongst the $K$ nearest neighbors of $x$ in the training set. 
To accelerate all KNN computations, we utilize the \texttt{Annoy} approximate neighbors library \cite{annoy}.

Finally we consider another OOD detection method that solely depends on learned image embeddings. Following \citet{luan2021out}, we apply an \textbf{Isolation Forest} to the image embeddings from our trained classfication network.

\section{Results} 

Tables \ref{tab:swin_auroc} and \ref{tab:resnet_auroc} list the AUROC achieved by each OOD detection method  when utilized with the Swin Transformer and ResNet-50 model on each in/out-of distribution dataset pair.
With the Swin Transformer, the KNN Distance overall tends to be the best method for OOD image detection. 
With ResNet-50, the KNN Distance also outperforms most of the other OOD methods, with the exception of RMD. RMD however performs significantly worse than KNN Distance and other OOD methods when the in/out-of distribution datasets are roman-numeral vs.\ mnist and the Swin Transformer model is used. 
\looseness=-1

Contrary to findings by \citet{hendrycks2016baseline, Vaze2022good}, MSP is not a very competitive baseline in our benchmark. 
For example, Entropy appears to be a universally superior drop-in replacement for MSP (both are simple functions of the model predictions). Two images may have the same MSP but greatly differing entropies if say one image is predicted to belong to one of two classes with equal confidence while the other image is predicted to belong to one class with 50\% confidence but equal likelihood for all of the remaining classes. Intuitively, the former image is less likely to be OOD vs.\ simply being a hard-to-classify instance for which two class labels might reasonably apply.

The Entropy and MSP methods that solely depend on model predictions are reliably outperformed by OOD methods that leverage intermediate representations from the model instead. 
This agrees with the basic intuition that effective OOD detection requires more information about the image (as captured in the intermediate learned representations) than is available in the predictions alone. Our results suggest the surprising findings of \citet{hendrycks2016baseline,Vaze2022good} are indeed too fortuitous to hold universally, and we hesitate to recommend any OOD detection solely based on the predictions of an arbitrary classifier.

There is valuable information in the model  representations (i.e.\ embeddings) that should be leveraged for OOD detection. 
Since the model predictions are simply a deterministic function of these embeddings, they do not provide much additional value once the embeddings are being effectively utilized (compare KNN Distance vs.\ KNN DistPred vs.\ KNN Prediction in 
Tables \ref{tab:swin_auroc} and \ref{tab:resnet_auroc}).

\begin{figure*}[b!] \centering
\vspace*{1mm}
\begin{tabular}{cc}
    \includegraphics[width=0.49\textwidth]{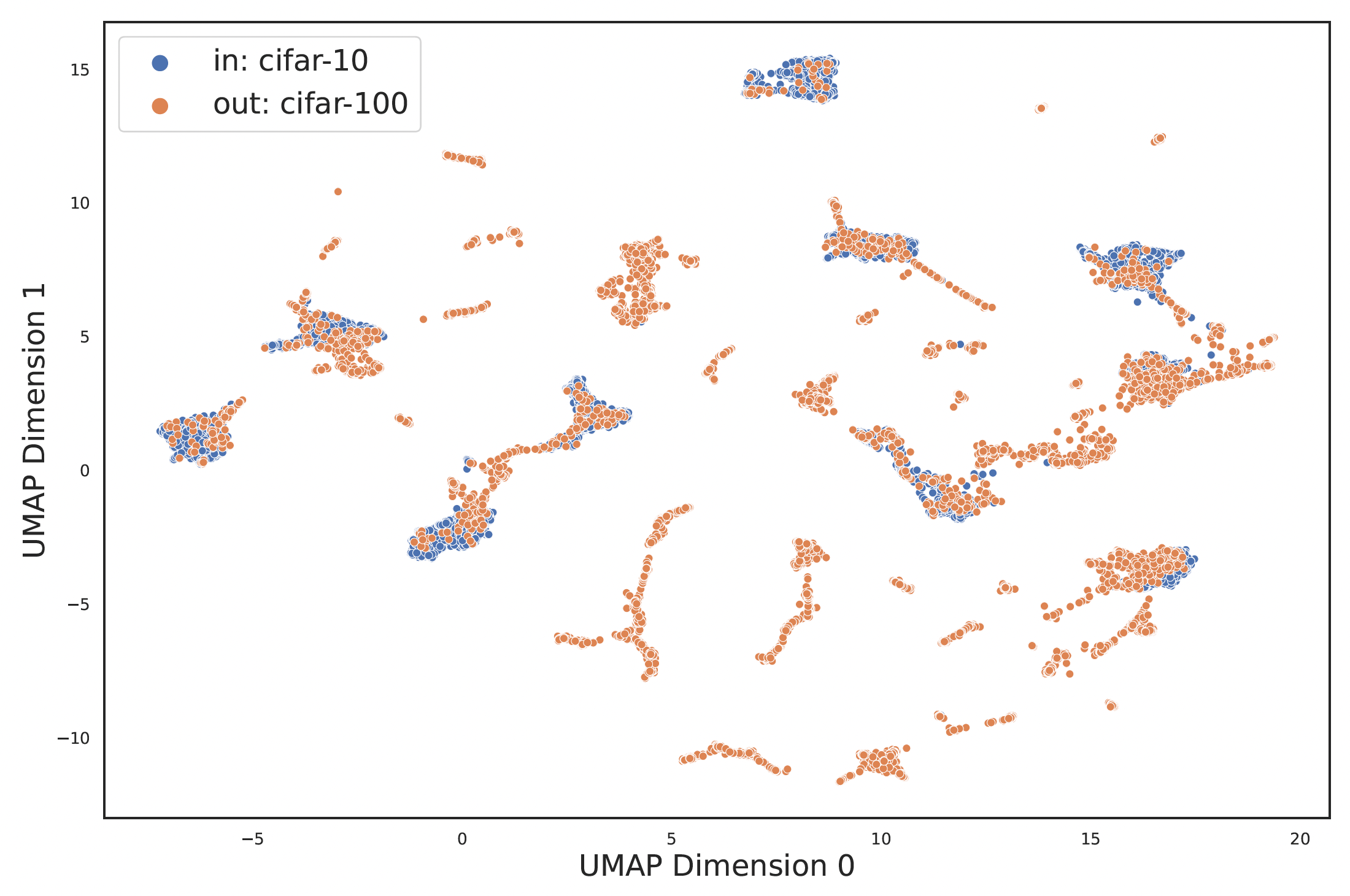}
&
\includegraphics[width=0.49\textwidth]{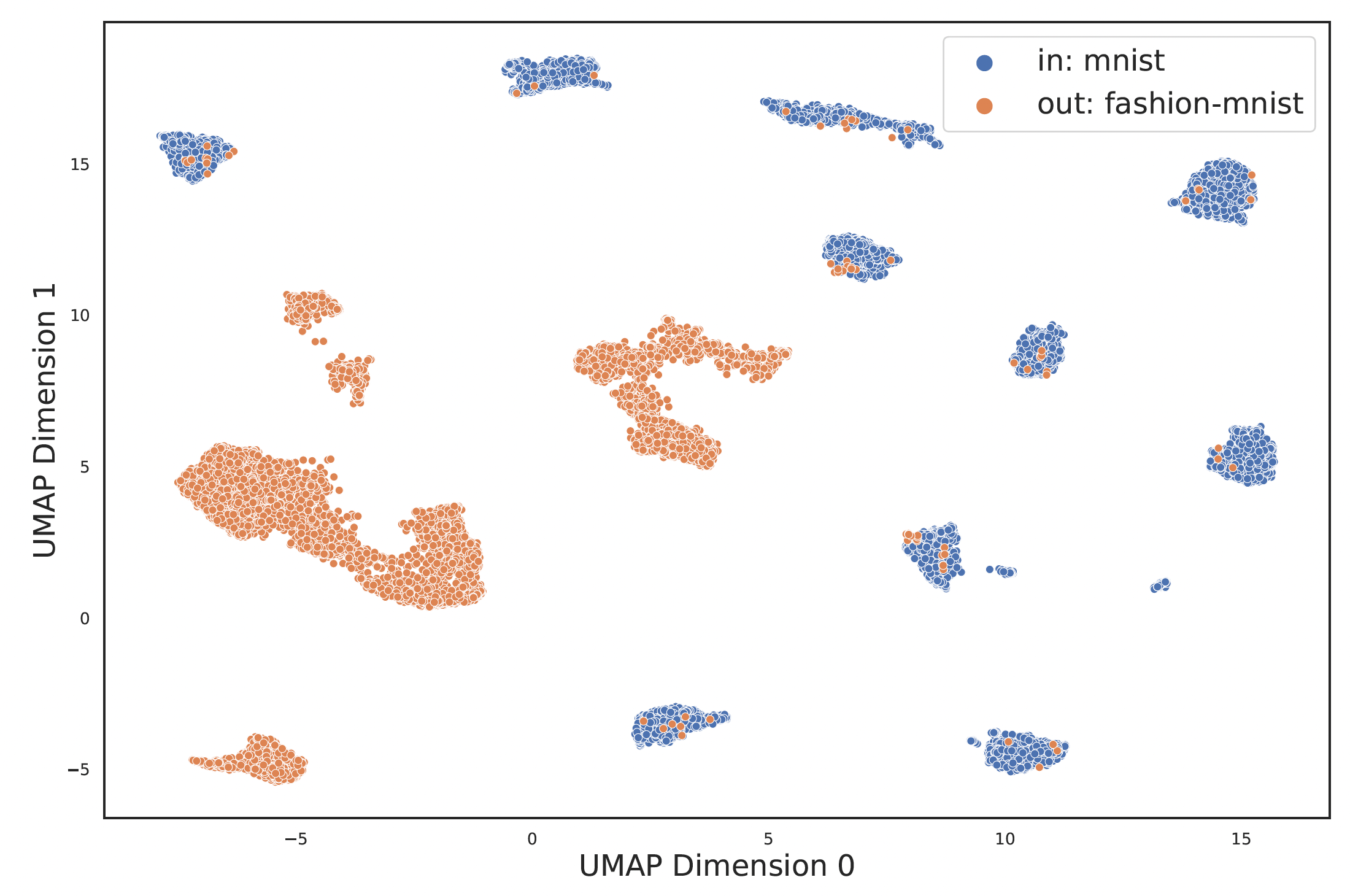}
    \\
  \hspace*{0mm}  \textbf{(A)} cifar-10 (in-distribution) vs.\ cifar-100 (OOD)
  & 
  \hspace*{0mm}  \textbf{(B)} mnist (in-distribution) vs.\ fashion-mnist (OOD)
\end{tabular}
    \caption{UMAP of learned image embeddings from the Swin Transformer model fit to the training set of the in-distribution dataset. For a particular  pairing of datasets to consider   in-distribution vs.\ OOD, each depicted dot is an image from the test set of either the in-distribution or out-of-distribution dataset. 
    }
    \label{fig:umap}
\end{figure*}

\begin{figure*}[h!] \centering
\begin{tabular}{cc}
    \includegraphics[width=0.49\textwidth]{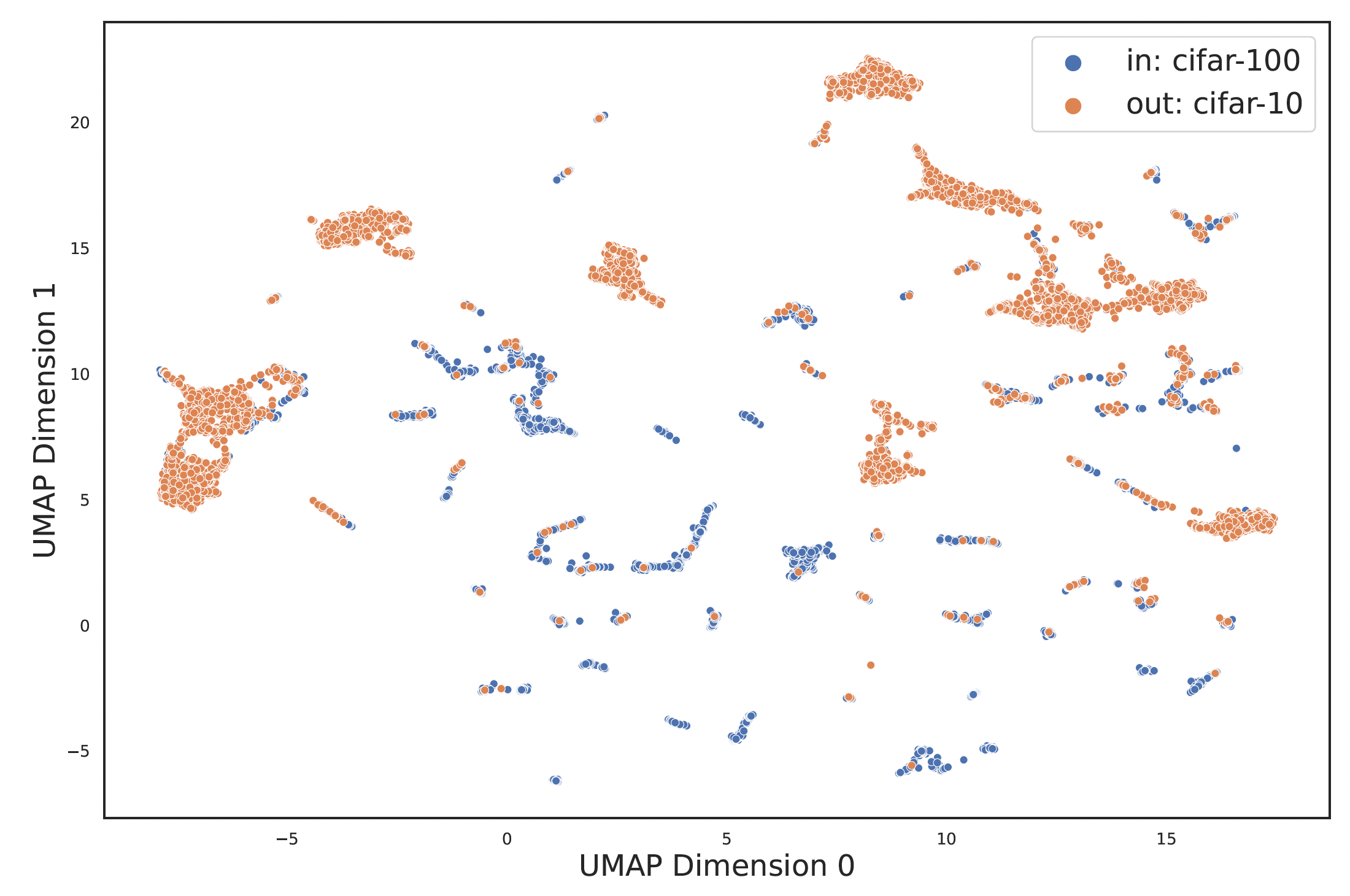} 
&
    \includegraphics[width=0.49\textwidth]{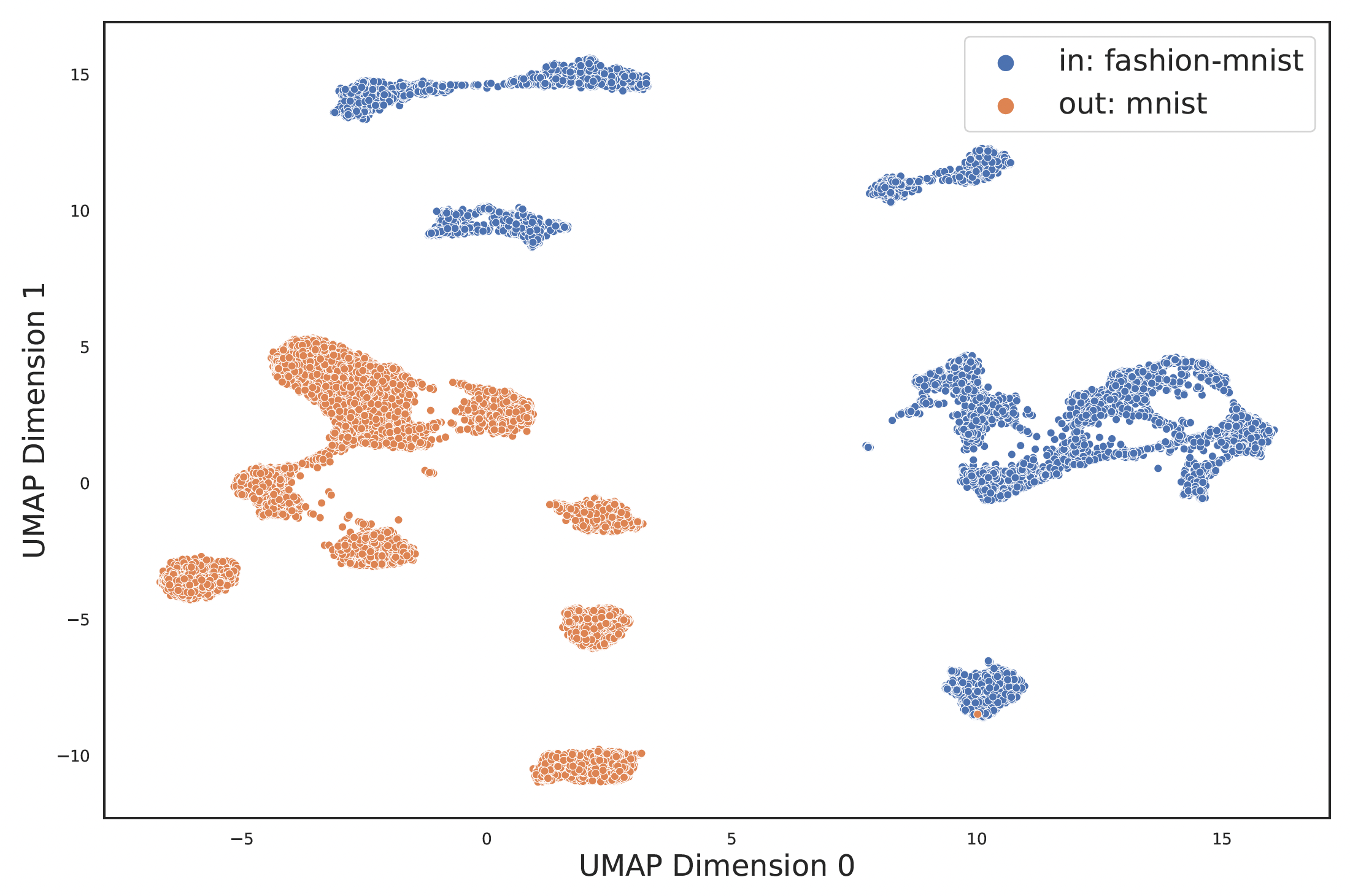} 
    \\
  \hspace*{0mm}  \textbf{(A)} cifar-100 (in-distribution) vs.\ cifar-10 (OOD)
  & 
  \hspace*{0mm}  \textbf{(B)} fashion-mnist (in-distribution) vs.\ mnist (OOD) 
  \\[0.5em]
  \includegraphics[width=0.49\textwidth]{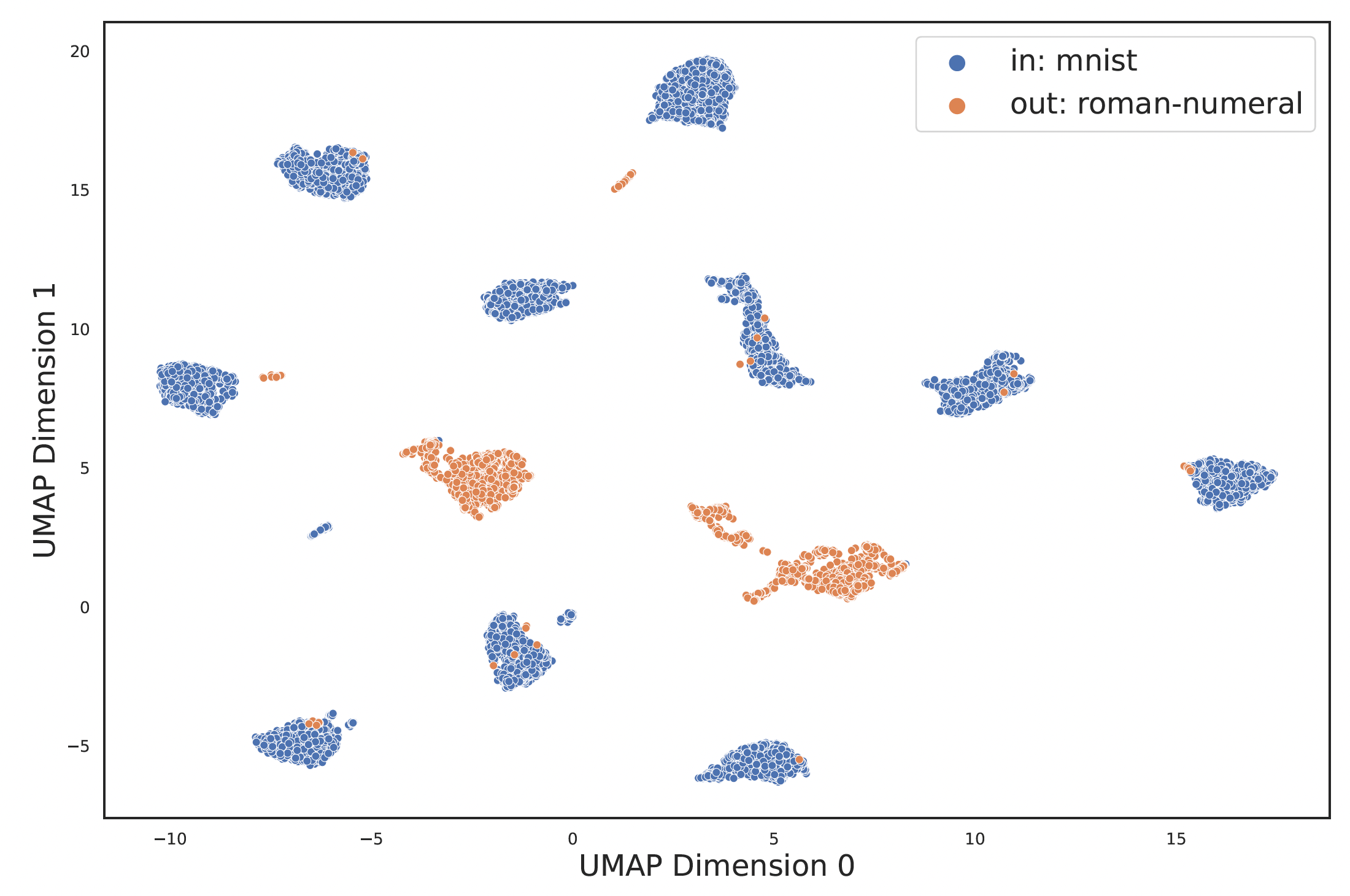}
&
    \includegraphics[width=0.49\textwidth]{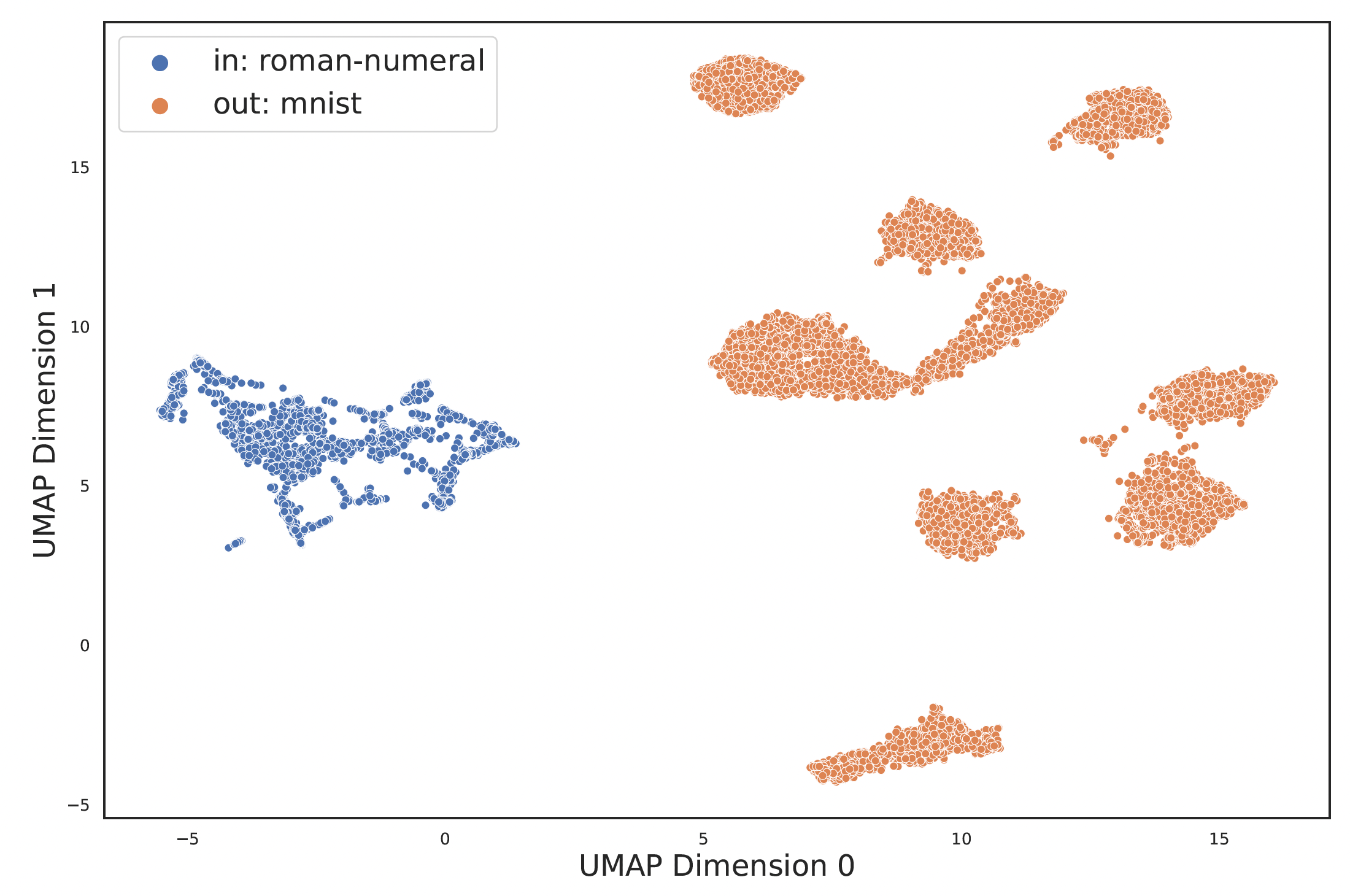} 
    \\
  \hspace*{0mm}  \textbf{(C)} mnist (in-distribution) vs.\ roman-numeral (OOD)
  & 
  \hspace*{0mm}  \textbf{(D)} roman-numeral (in-distribution) vs.\ mnist (OOD)
  \\
\end{tabular}
    \caption{UMAP of learned image embeddings from the Swin Transformer model fit to the training set of the in-distribution dataset. For a particular  pairing of datasets to consider   in-distribution vs.\ OOD, each depicted dot is an image from the test set of either the in-distribution or out-of-distribution dataset. Shown here are the remaining dataset pairs missing from Figure \ref{fig:umap}.
    }
    \label{fig:umap2}
\end{figure*}

However the class labels $y$ do provide useful auxiliary information. Even though the effective KNN Distance method is solely a function of the image embeddings, recall these embeddings are obtained from a classifier trained using the labels. 
Tables \ref{fig:umap} and \ref{fig:umap2} depict the learned embeddings of test images, revealing distinct  clustering of the in-distribution data based on its class labels. While the embeddings of OOD data do not exhibit similar cluster structure, it appears the label-informed geometry of the in-distribution embeddings suffices to ensure that the KNN Distance can adequately distinguish them from the OOD embeddings. 
Note also that the Mahalanobis and RMD methods depend on the embeddings and class labels -- with the latter method being particularly effective for OOD detection with ResNet-50 in cifar-100 vs. cifar-10 and fashion-mnist vs.\ mnist (Table \ref{tab:resnet_auroc}).

Tables \ref{tab:kvals_transformer} and \ref{tab:kvals_resnet} show the performance of the KNN-based methods for OOD detection is quite stable over the choice of $K$. We also experimented with more adaptive schemes to choose the $K$ that maximizes KNN-classification accuracy, but did not observe noticeable benefits.

\textbf{Discussion.} \ 
Our study suggests the KNN Distance deserves greater consideration for practical out-of-distribution image detection pipelines. This approach is: straightforward to implement, conceptually well-grounded, 
highly interpretable, amenable to various computational shortcuts/approximations, and capable of leveraging any classification  model. 

\clearpage

\begin{table*}[h!]

\caption{AUROC of OOD detection using Swin Transformer model. For KNN methods, the results shown here use $K=10$. 
\\
Results that are substantially better than the others on the same dataset are highlighted in bold.}
\label{tab:swin_auroc}

{\scriptsize
\begin{tabular}{llcccccccc}
\toprule
In Distribution & Out of Distribution &    MSP &  Entropy &  Mahalanobis &    RMD &  Isolation Forest &  KNN Distance &  KNN DistPred &  KNN Prediction  \\
\midrule
       cifar-10 &           cifar-100 & 0.9689 &   0.9690 &       0.9895 & 0.9851 &            0.9245 &               \textbf{0.9917} &                             \textbf{0.9916} &                  0.9700 \\
      cifar-100 &            cifar-10 & 0.9470 &   \textbf{0.9558} &       0.9128 & 0.9485 &            0.5907 &               \textbf{0.9571} &                             \textbf{0.9570} &                  0.9549 \\
          mnist &       roman-numeral & 0.9924 &   0.9946 &       0.9997 & 0.9952 &            0.9914 &               0.9989 &                             0.9989 &                  0.9952 \\
  roman-numeral &               mnist & 0.9090 &   0.9299 &       0.9953 & 0.8753 &            0.7521 &               \textbf{0.9957} &                             \textbf{0.9958} &                  0.9362 \\
          mnist &       fashion-mnist & 0.9945 &   0.9956 &       0.9998 & 0.9970 &            0.9853 &               0.9997 &                             0.9997 &                  0.9964 \\
  fashion-mnist &               mnist & 0.9137 &   0.9377 &       \textbf{0.9982} & \textbf{0.9944} &            0.9734 &               \textbf{0.9945} &                             \textbf{0.9944} &                  0.9434 \\
\bottomrule
\end{tabular}
}
\end{table*}

\begin{table*}[h!]

\caption{AUROC of OOD detection using ResNet-50 model. For KNN methods, the results shown here use $K=10$. }
\label{tab:resnet_auroc}

{\scriptsize
\begin{tabular}{llcccccccc}
\toprule
In Distribution & Out of Distribution &    MSP &  Entropy &  Mahalanobis &    RMD &  Isolation Forest &  KNN Distance &  KNN DistPred &  KNN Prediction  \\
\midrule
       cifar-10 &           cifar-100 & 0.9268 &   \textbf{0.9318} &       0.8698 & \textbf{0.9315} &            0.6385 &               \textbf{0.9390} &                             \textbf{0.9388} &                  \textbf{0.9311} \\
      cifar-100 &            cifar-10 & 0.7864 &   0.8001 &       0.6988 & \textbf{0.8344} &            0.6120 &               0.8041 &                             0.8052 &                  0.8036 \\
          mnist &       roman-numeral & 0.9843 &   0.9884 &       \textbf{1.0000} & 0.9981 &            0.9559 &               \textbf{1.0000} &                             \textbf{1.0000} &                  0.9896 \\
  roman-numeral &               mnist & 0.8750 &   0.9239 &       \textbf{0.9999} & 0.9994 &            0.7446 &               \textbf{1.0000} &                             \textbf{1.0000} &                  0.9110 \\
          mnist &       fashion-mnist & 0.9826 &   0.9865 &       \textbf{0.9999} & \textbf{0.9952} &            0.6717 &               \textbf{0.9997} &                             \textbf{0.9997} &                  0.9882 \\
  fashion-mnist &               mnist & 0.7369 &   0.7400 &       \textbf{0.9965} & \textbf{0.9941} &            0.8903 &               0.9450 &                             0.9457 &                  0.7676 \\
\bottomrule
\end{tabular}
}

\end{table*}

\setlength{\tabcolsep}{4pt} 

\begin{table*}[h!]

\caption{AUROC of OOD detection using Swin Transformer model and various KNN methods with different values of $K$.}
\label{tab:kvals_transformer}

{\scriptsize
\begin{tabular}{llcccccccccccc}
\toprule
In & OOD &  Distance & Distance &  Distance &  Distance  &  DistPred &  DistPred  &  DistPred &  DistPred &  Prediction &  Prediction &  Prediction &  Prediction 
\\
Distribution & & K=5 & K=10 & K=15 & K=100 & K=5 & K=10 & K=15 & K=100 & K=5 & K=10 & K=15 & K=100
\\ 
\midrule
       cifar-10 &           cifar-100 &              0.9918 &               0.9917 &               0.9916 &                0.9902 &                            0.9917 &                             0.9916 &                             0.9915 &                              0.9901 &                 0.9701 &                  0.9700 &                  0.9699 &                   0.9697 \\
      cifar-100 &            cifar-10 &              0.9562 &               0.9571 &               0.9574 &                0.9566 &                            0.9562 &                             0.9570 &                             0.9573 &                              0.9565 &                 0.9548 &                  0.9549 &                  0.9548 &                   0.9541 \\
          mnist &       roman  &              0.9989 &               0.9989 &               0.9988 &                0.9984 &                            0.9989 &                             0.9989 &                             0.9988 &                              0.9984 &                 0.9951 &                  0.9952 &                  0.9952 &                   0.9952 \\
  roman &               mnist &              0.9959 &               0.9957 &               0.9955 &                0.9934 &                            0.9959 &                             0.9958 &                             0.9955 &                              0.9935 &                 0.9299 &                  0.9362 &                  0.9396 &                   0.9408 \\
          mnist &       fashion &              0.9997 &               0.9997 &               0.9997 &                0.9995 &                            0.9997 &                             0.9997 &                             0.9997 &                              0.9995 &                 0.9964 &                  0.9964 &                  0.9964 &                   0.9963 \\
  fashion &               mnist &              0.9950 &               0.9945 &               0.9942 &                0.9915 &                            0.9948 &                             0.9944 &                             0.9940 &                              0.9913 &                 0.9432 &                  0.9434 &                  0.9433 &                   0.9411 \\
\bottomrule
\end{tabular}
}

\end{table*}

\begin{table*}[h!]

\caption{AUROC of OOD detection using ResNet-50 model and various KNN methods with different values of $K$.}
\label{tab:kvals_resnet}

{\scriptsize
\begin{tabular}{llcccccccccccc}
\toprule
In & OOD &  Distance & Distance &  Distance &  Distance  &  DistPred &  DistPred  &  DistPred &  DistPred &  Prediction &  Prediction &  Prediction &  Prediction 
\\
Distribution & & K=5 & K=10 & K=15 & K=100 & K=5 & K=10 & K=15 & K=100 & K=5 & K=10 & K=15 & K=100
\\ 
\midrule
       cifar-10 &           cifar-100 &              0.9388 &               0.9390 &               0.9387 &                0.9349 &                            0.9387 &                             0.9388 &                             0.9386 &                              0.9345 &                 0.9313 &                  0.9311 &                  0.9307 &                   0.9268 \\
      cifar-100 &            cifar-10 &              0.8086 &               0.8041 &               0.8005 &                0.7689 &                            0.8099 &                             0.8052 &                             0.8017 &                              0.7715 &                 0.8050 &                  0.8036 &                  0.8016 &                   0.7821 \\
          mnist &       roman &              1.0000 &               1.0000 &               1.0000 &                1.0000 &                            1.0000 &                             1.0000 &                             1.0000 &                              1.0000 &                 0.9896 &                  0.9896 &                  0.9895 &                   0.9884 \\
  roman &               mnist &              0.9999 &               1.0000 &               1.0000 &                0.9997 &                            0.9999 &                             1.0000 &                             1.0000 &                              0.9997 &                 0.9012 &                  0.9110 &                  0.9175 &                   0.9450 \\
          mnist &       fashion &              0.9997 &               0.9997 &               0.9997 &                0.9996 &                            0.9997 &                             0.9997 &                             0.9997 &                              0.9995 &                 0.9883 &                  0.9882 &                  0.9879 &                   0.9869 \\
  fashion &               mnist &              0.9506 &               0.9450 &               0.9409 &                0.9106 &                            0.9516 &                             0.9457 &                             0.9415 &                              0.9106 &                 0.7719 &                  0.7676 &                  0.7651 &                   0.7486 \\
\bottomrule
\end{tabular}
}

\end{table*}

\clearpage

\bibliographystyle{icml2022}
\bibliography{ood_references}



\clearpage \newpage
\beginsupplement
\onecolumn
\appendix

\def\toptitlebar{\hrule height1pt \vskip .2in} 
\def\bottomtitlebar{\vskip .22in \hrule height1pt \vskip .3in} 

\thispagestyle{empty}

\setcounter{page}{1}
\pagenumbering{arabic}
\setlength{\footskip}{20pt}  
\vspace*{-3.5mm}
\begin{center}
\toptitlebar
{\Large \bf Supplementary Material 
}
\bottomtitlebar
\end{center}


\vspace*{1mm} 

\vspace*{5mm}
\begin{table*}[b!]

\caption{Test accuracy for predicting class labels of in-distribution data achieved by each model.}
\label{tab:acc}
\vspace*{1em}

\begin{center}
\begin{tabular}{lrr}
\toprule
      Dataset &  Swin Transformer &  ResNet-50 \\
\midrule
     cifar-10 &            0.9887 &     0.9665 \\
    cifar-100 &            0.9286 &     0.8288 \\
roman-numeral &            0.7967 &     0.7686 \\
        mnist &            0.9914 &     0.9890 \\
fashion-mnist &            0.9489 &     0.9424 \\
\bottomrule
\end{tabular}
\end{center}
\end{table*}

\end{document}